\documentclass[letterpaper]{article}

\usepackage{aaai2027}

\usepackage[hyphens]{url}
\usepackage{graphicx}
\usepackage{natbib}
\usepackage{caption}
\usepackage{algorithm}
\usepackage{algorithmic}

\usepackage{newfloat}
\usepackage{listings}
\DeclareCaptionStyle{ruled}{labelfont=normalfont,labelsep=colon,strut=off}
\floatstyle{ruled}
\newfloat{listing}{tb}{lst}{}
\floatname{listing}{Listing}
\usepackage{amsthm}

\newtheorem*{theorem*}{Theorem}

\newtheorem{proposition}{Proposition}

\usepackage{booktabs}

\usepackage{graphicx}
\usepackage{subfig}
\usepackage{amsmath}
\usepackage{multirow}
\usepackage[table]{xcolor}
\usepackage{amssymb}

\definecolor{tabhighlight}{RGB}{245,245,245}
\definecolor{lightgreen}{RGB}{230,230,230}

\newcommand{\moiraismall}{Moirai-small}
\newcommand{\moiraibase}{Moirai-base}

\title{Ground-Truth Neighborhood Regularization for Reinforcement Learning Post-Training of Time Series Foundation Models}
\author{
    Jianqi Zhang\textsuperscript{\rm 1},
    Xingyu Zhang\textsuperscript{\rm 1},
    Zeen Song\textsuperscript{\rm 1},
    Changwen Zheng\textsuperscript{\rm 1},
    Fanjiang Xu\textsuperscript{\rm 1},
    Wenwen Qiang\textsuperscript{\rm 1}
}

\affiliations{
    \textsuperscript{\rm 1}Institute of Software, Chinese Academy of Sciences, Beijing, China\\
    jluzhangjianqi@163.com,
    xingyuzhang@mails.ucas.ac.cn,
    zeensong@qq.com,\\
    changwen@iscas.ac.cn,
    fanjiang@iscas.ac.cn,
    qiangwenwen@iscas.ac.cn
}

\begin{document}

\nocopyright
\maketitle

\begin{center}
\textit{Preprint. Under review.}
\end{center}

\begin{abstract}

Time series forecasting (TSF) plays an important role in a wide range of real-world applications. Recently, time series foundation models (TSFMs), pretrained on large-scale datasets, have demonstrated strong generalization capabilities and emerged as an important paradigm for TSF. Reinforcement learning (RL) post-training has consequently attracted growing attention as a means of further improving their performance on downstream tasks. However, we find that, in certain forecast regions, RL post-training may gradually shift the output distributions of TSFMs away from the ground truth, thereby limiting their performance. We refer to this phenomenon as \textbf{suboptimal collapse}. Our analysis suggests that difficulty in initially sampling high-quality trajectories near the ground truth is an important contributing factor to suboptimal collapse. To address this issue, we propose Ground-Truth Neighborhood Regularization (GTN-R) for RL post-training of TSFMs. GTN-R uses the ground truth as a reference for locating high-quality regions and guides the model’s probability mass toward the ground-truth neighborhood. This increases the probability of sampling high-quality trajectories, mitigates suboptimal collapse, and improves performance. Moreover, GTN-R can be flexibly integrated into various RL methods for TSFMs. Extensive experiments show its effectiveness.
\end{abstract}

\section{Introduction}

Time series forecasting (TSF) is a fundamental task in a wide range of applications, including energy scheduling \cite{boussif2024improving,novo2022planning,lara2020temporal,deb2017review}, traffic management \cite{fang2023stwave+,li2022dmgan,wang2022st,shekhar2007adaptive}, and weather forecasting \cite{abhishek2012weather, karevan2020transductive,campbell2005weather,hamilton2007ski}. Recently, time series foundation models (TSFMs), pretrained on large-scale time series data, have demonstrated strong generalization across diverse scenarios \cite{woo2024unified}. However, the general capabilities acquired during pretraining may not fully adapt to specific downstream tasks, making post-training necessary to improve performance in target scenarios further \cite{qiao2025multi}. Recent studies \cite{qi2025timehf,li2026timerft} have shown that, compared with supervised fine-tuning (SFT), reinforcement learning (RL) post-training can more effectively unlock the forecasting capabilities of TSFMs on downstream tasks, thereby attracting increasing attention.

\begin{figure}[t]
    \centering
    \includegraphics[width=\columnwidth]{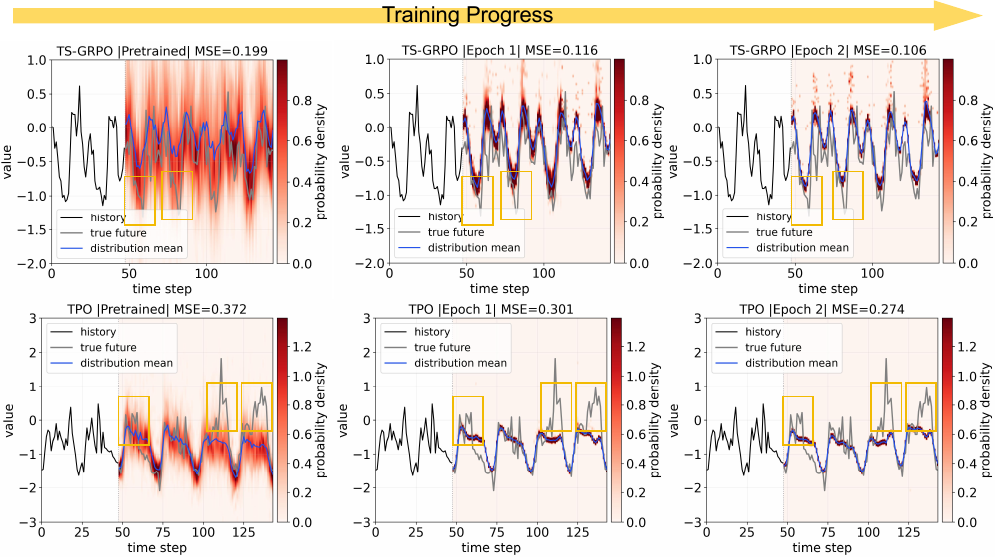}
    \caption{Illustration of suboptimal collapse during RL post-training of Moirai-small. Each row shows the evolving predictive distribution for a training sample, with the RL method, processed epoch, and MSE reported above each panel. The TSFM used is Moirai-small.As training progresses, distribution within the yellow boxes shifts toward regions far from the ground truth, indicating suboptimal collapse. More results are in Appendix “More Visualization.”
    }
    \label{collapse}
\end{figure}

Although RL post-training improves the forecasting performance of TSFMs, we identify a notable issue during training. Specifically, ideally, as training progresses, the model’s predictive distribution should gradually concentrate around the ground truth. However, we observe that, in many cases, although the overall MSE decreases, the predictive distribution in some regions (yellow boxes) gradually concentrates outside the ground-truth neighborhood, as shown in Fig.\ref{collapse}. We term this phenomenon \textbf{suboptimal collapse}. It may trap subsequent optimization in low-quality regions far from the ground truth and limit forecasting performance. From an optimization perspective, suboptimal collapse may arise because the model initially struggles to sample high-quality trajectories near the ground truth in certain regions. In this case, the relatively better trajectories within a sampled group are often located in low-quality regions far from the ground truth. Since RL optimizes the model by increasing the probabilities of relatively better trajectories \cite{qi2025timehf,li2026timerft}, updates in the above sampled group further widen the probability gap between the low-quality region and the ground-truth neighborhood. This makes the model more likely to sample trajectories from the low-quality region and, consequently, makes the relatively better trajectories in subsequent sampled groups more likely to be in the low-quality region. Repeated occurrences of this process form a detrimental self-reinforcing cycle, which may eventually drive the probability of the ground-truth neighborhood toward zero, thereby leading to suboptimal collapse. In Sections “Theoretical Analysis” and “Empirical Analysis”, we further validate this intuitive conjecture from theoretical and empirical perspectives, respectively.

An intuitive solution is to directly include the ground-truth trajectory in the sampled group. However, because the ground-truth participates in the updates as a fixed sample with the highest reward, the model may instead learn to imitate it in a supervised manner, causing RL post-training to gradually degenerate into a process resembling behavior cloning \cite{yan2026s}. This may weaken the model’s exploration capability and further impair its generalization ability \cite{wang2025benefits,weltevrede2024explore,chu2025sft}. Our experiments in Section “Empirical Analysis” further support this argument: although this approach mitigates suboptimal collapse on the training data, it undermines the model’s exploration capability and generalization performance. Therefore, although the ground-truth trajectory can serve as a reference for high-quality trajectories, it should not participate in the updates as a fixed high-reward sample.

Based on the above analysis, we propose Ground-Truth Neighborhood Regularization (GTN-R), a regularization method for RL post-training of TSFMs. Instead of directly including the ground truth in the updates as a fixed high-reward sample, GTN-R uses it only as a reference for high-quality trajectories, guiding the model to increase the probability of sampling high-quality trajectories. Consequently, GTN-R mitigates suboptimal collapse while preserving the model’s autonomous exploration capability, thereby further improving performance. Specifically, GTN-R introduces two distributional constraints into the original RL objective: the neighborhood probability mass constraint and the within-neighborhood uniformity constraint. The former maximizes the probability mass assigned by the predictive distribution to the ground-truth neighborhood, thereby increasing the probability of sampling high-quality trajectories and mitigating suboptimal collapse. The latter encourages the distribution within this neighborhood to approach a uniform distribution, preventing probability mass from concentrating at only a few locations and thus preserving the model’s exploration ability. We integrate GTN-R into multiple RL methods applicable to TSFMs. Experiments across multiple TSFMs and datasets show that GTN-R effectively mitigates suboptimal collapse and improves performance.

Our contributions are as follows: \textbf{1)} We identify \textbf{suboptimal collapse} in RL post-training of TSFMs and, through theoretical analysis and empirical studies, investigate its mechanism and negative impact on forecasting performance. \textbf{2)} To address this issue, we propose GTN-R, which mitigates suboptimal collapse by increasing the probability mass assigned to the ground-truth neighborhood while encouraging a near-uniform distribution within it, thereby improving forecasting performance. \textbf{3)} We validate the effectiveness of GTN-R across multiple RL methods for TSFMs.

\section{Related Work}

Recently, Time Series Foundation Models (TSFMs) have advanced rapidly \cite{miller2024survey,kottapalli2025foundation,ye2026empowering,liu2025empowering}. Some TSFMs, such as Moment \cite{goswami2024moment}, UniTS \cite{gao2024units}, TimesFM \cite{das2024decoder}, and Timer \cite{liu2024timer}, produce deterministic forecasts. However, stochastic disturbances, observation noise, and environmental changes make future time series inherently uncertain \cite{clark2004population,yoon2022robust}. Deterministic predictions cannot capture multiple possible futures or quantify predictive confidence, thereby limiting their ability to support uncertainty-aware decision-making in real-world \cite{liu2024moirai}. This has motivated probabilistic TSFMs (Moirai \cite{woo2024unified}, Moirai-MoE \cite{liu2024moirai}, and Toto \cite{cohen2024toto}), which produced the distribution of future series.

For downstream-task post-training, the research community initially relies on supervised fine-tuning (SFT) to optimize TSFMs \cite{qiao2025multi}. However, \cite{qiao2025multi,qi2025timehf,li2026timerft} shows that naive SFT cannot fully unlock TSFMs' forecasting potential. Recent studies \cite{qi2025timehf,li2026timerft}, therefore, apply RL methods, such as RLHF \cite{ouyang2022training} and GRPO \cite{shao2024deepseekmath}, to TSFM post-training and achieve better performance than SFT. However, we find that directly applying RL may lead to suboptimal collapse, as shown in Fig.\ref{collapse}. This paper aims to mitigate suboptimal collapse in RL post-training of TSFMs and further improve forecasting performance.

\section{Preliminaries}
\label{preliminaries}
\subsection{Time Series Forecasting (TSF).}
In TSF, given a historical series $X\in \mathbb{R}^{T_h \times N}$, where $T_h$ is the input length, $N$ is the number of variables, the task is to predict the series of the next $T_f$ time steps $Y\in\mathbb{R}^{T_f \times N}$.

\subsection{Probabilistic Forecasting and Sampling in TSFMs}

In this paper, we focus on probabilistic TSFMs because they explicitly model predictive distributions of future series, from which RL can conveniently sample trajectories for update. Probabilistic TSFMs mainly adopt two forecasting paradigms: one-shot forecasting and token-by-token forecasting \cite{liu2026pre}. We next describe how these two get future predictive distributions and sample trajectories.

We first introduce the one-shot forecasting paradigm, which is adopted by Moirai \cite{woo2024unified}. Given a historical series $X$, the model simultaneously predicts the distributions of all future time points through a single forward pass, yielding the set of all predictive distributions:
\begin{equation}
    f_\theta(X)=\{P(\hat{y}_1\mid X),...,P(\hat{y}_{T_f}\mid X)\}=\mathcal{P},
\end{equation}
where $f_\theta$ denotes the forecasting function and $P(\hat{y}_t\mid X)$ the predictive distribution of the $t$-th point of $\hat{Y}$. We then sample from the point-wise distributions in $\mathcal{P}$ to construct multiple trajectories for training:
\begin{equation}
    \{\hat{Y}^i\}_{i=1}^n\sim \mathcal{P}.
\end{equation}

We next introduce the token-by-token forecasting paradigm, which is adopted by Moirai-MoE \cite{liu2024moirai} and Toto \cite{cohen2024toto}. Given a historical series $X$, the future series $Y$ is divided into contiguous segments of equal length $\{Y_{s1}, Y_{s2}, \ldots\}$, each corresponding to one token. The model first predicts the distribution of each point in $Y_{s1}$:
\begin{equation}
    f_\theta(X)=\{P(\hat{y}_{s1,1}\mid X),...,P(\hat{y}_{s1,T_p}\mid X)\}=\mathcal{P}_{s1},
\end{equation}
where $T_p$ is the length of each patch, $P(\hat{y}_{s1,t}\mid X)$ is the $t$-th-point predictive distribution of $\hat{Y}_{s1}$. We then sample from the point-wise distributions in $\mathcal{P}_{s1}$ to construct multiple segment-level trajectories:
\begin{equation}
    \{\hat{Y}_{s1}^i\}_{i=1}^n\sim \mathcal{P}_{s1}.
\end{equation}
Next, for each $\hat{Y}_{s1}^i$, we concatenate it with the existing series to form a new series $(X,\hat{Y}_{s1}^i)$. Then it is fed into the model to predict the distribution of each point of the next segment, $\mathcal{P}_{s2}^i$=$\{P(\hat{y}_{s2,t}\mid X,\hat{Y}_{s1}^i)\}_{t=1}^{T_p}$. Next, we sample only one segment-level trajectory from $\mathcal{P}_{s2}^i$ as the continuation of $\hat{Y}_{s1}^i$. This process is repeated until each trajectory reaches the required prediction length.

\begin{figure*}[htbp]
    \centering
    \subfloat[]{\includegraphics[width=0.33\textwidth]{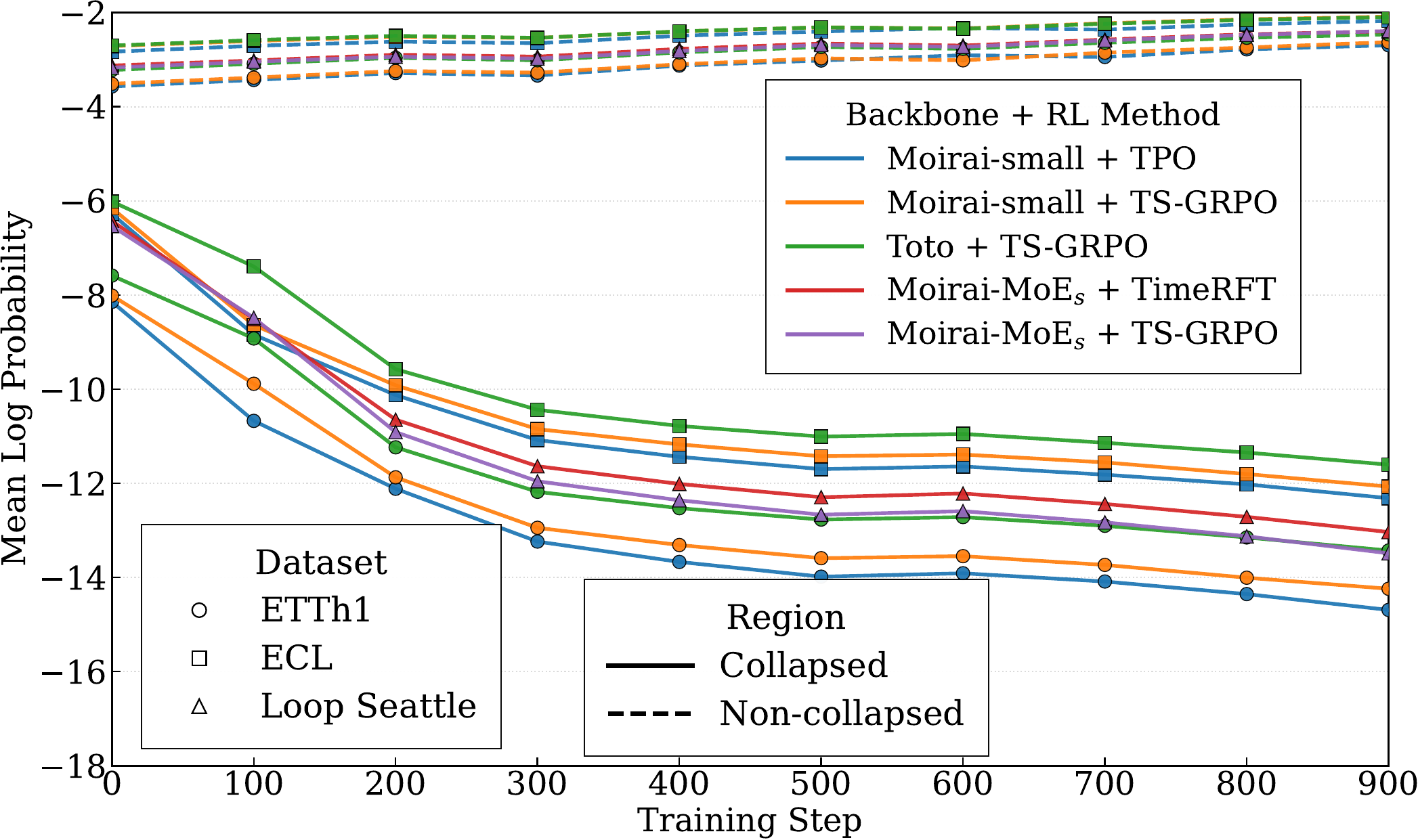}
    \label{fig_motivation1}}
    \hfil
    \subfloat[]{\includegraphics[width=0.33\textwidth]{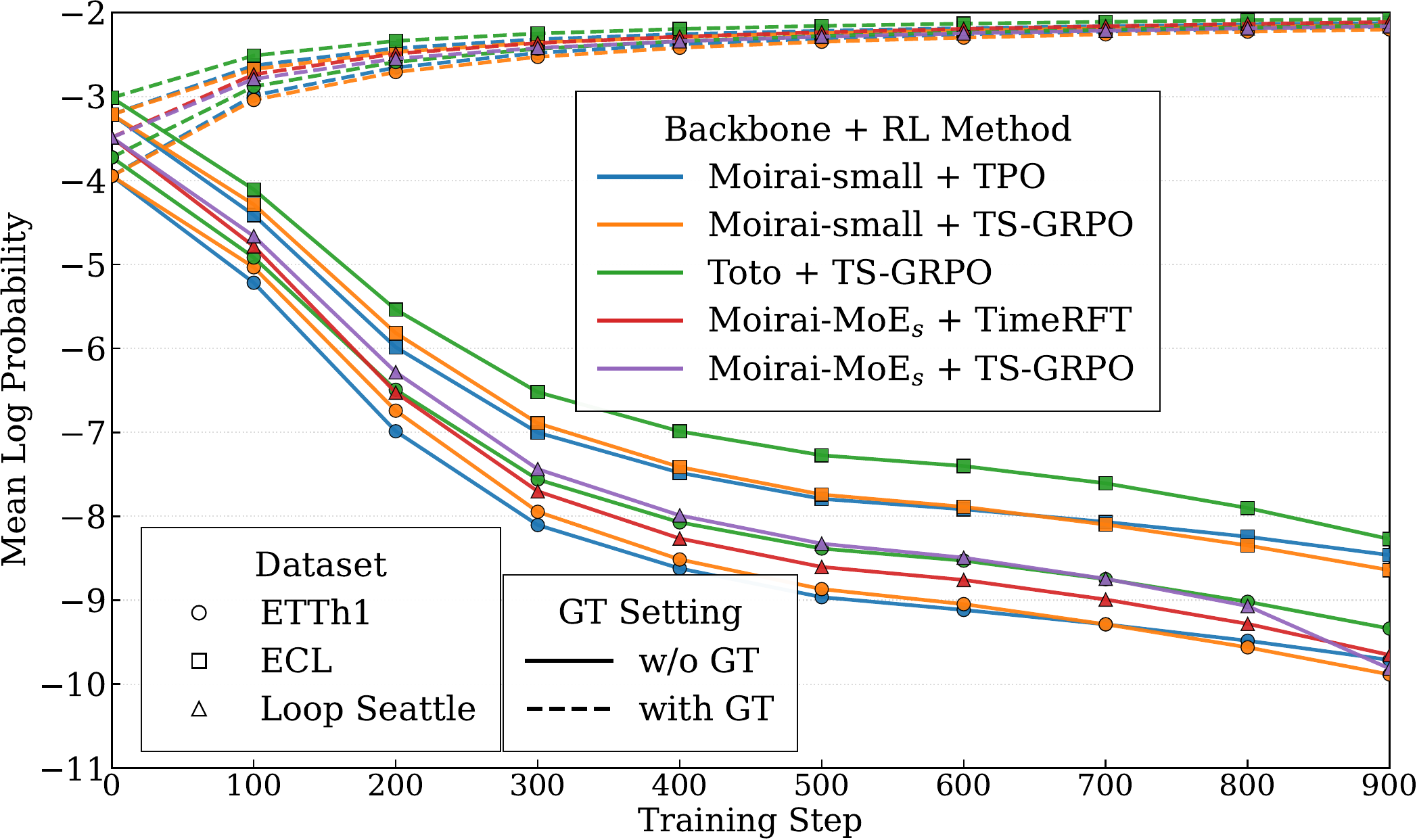}
    \label{fig_motivation2}}
    \hfil
    \subfloat[]{\includegraphics[width=0.33\textwidth]{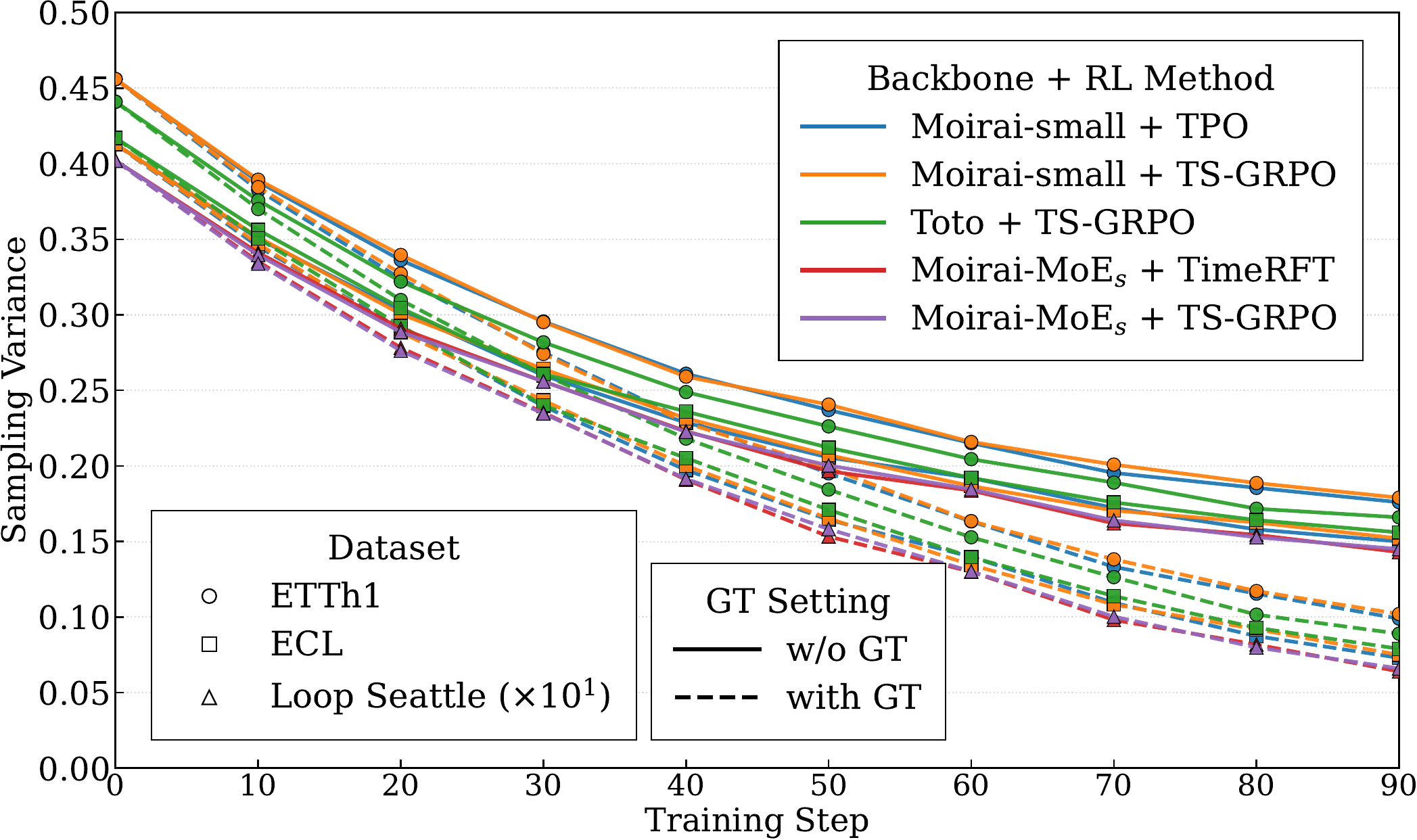}
    \label{fig_motivation3}}
    \hfil
    \caption{(a) Mean log probability assigned to the ground-truth neighborhoods of collapsed and non-collapsed points during post-training. Mean ground-truth-neighborhood log probability (b) and sampling variance (c) with and without including the ground-truth trajectory. Train step denotes the number of processed batches. In (c), $\times 10^1$ indicates that the actual variances of the dataset are 10 times the plotted values.
    }
    \label{fig_abla}
\end{figure*}

\begin{figure}[htbp]
    \centering
    \includegraphics[width=0.75\linewidth]{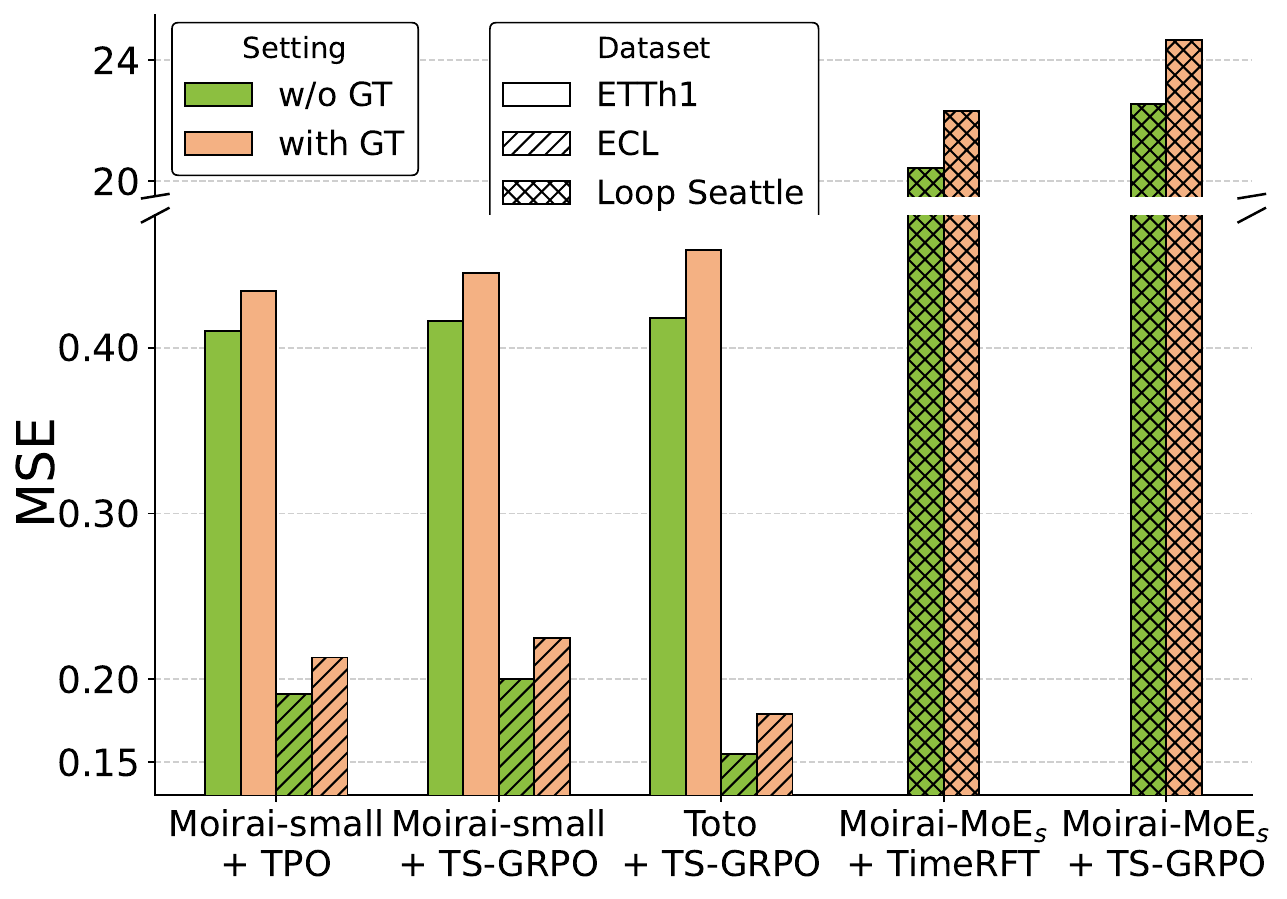}
    \caption{Test-set forecasting performance with and without including the ground-truth trajectory in the sampled group.}
    \label{fig_motivation4}
\end{figure}

\section{Theoretical Analysis}
\label{sec:theoretical_analysis}

In this section, we theoretically analyze a possible mechanism that may lead to suboptimal collapse. For a given future time point $t$ and its predictive context $\mathcal{C}_t$, let $\mathcal{G}=\mathcal{N}_r(y_t)$ be the ground-truth neighborhood, and let $\mathcal{S}$ be a low-quality region such that $\mathcal{G}\cap\mathcal{S}=\varnothing$. Their union is not required to cover the entire point-wise prediction space. Let $p_{\theta_k,t}(\cdot\mid\mathcal{C}_t)$ be the model's predictive probability at iteration $k$. We define:
\begin{equation}
p_k=p_{\theta_k,t}(\mathcal{G}\mid\mathcal{C}_t),
\qquad
s_k=p_{\theta_k,t}(\mathcal{S}\mid\mathcal{C}_t).
\end{equation}
Suppose that $M$ trajectories are independently sampled at each iteration, and let $\hat{y}_t^i$ denote the prediction at time point $t$ in the $i$-th trajectory. Let $E_k$ denote the event that at least one sampled prediction lies in $\mathcal{G}$, i.e., $\hat{y}_t^i\in\mathcal{G}$ for some $i$, and let $E_k^c$ denote its complement. We further let $A_k$ denote the event that $E_k^c$ occurs and the predictions at time point $t$ in the positively reinforced trajectories (i.e., the better trajectories among the $M$ samples) lie in $\mathcal{S}$.

Our analysis is based on the following conditions.

\textbf{Initial sampling difficulty.}
The probability $p_k$ assigned to the ground-truth neighborhood at time point $t$ is low.

\textbf{Low-quality-dominance update.}
When no sampled prediction at time point $t$ lies in $\mathcal{G}$, the values of the points of better trajectories that dominate the update are likely to lie in a certain low-quality region $\mathcal{S}$. Formally, there exists a sufficiently large $\rho\in(0,1)$ such that
\begin{equation}
    p(A_k\mid\mathcal{F}_k,E_k^c)\geq\rho,
    \label{eq:suboptimal_update_dominance}
\end{equation}
where $\mathcal{F}_k$ denotes the training history before iteration $k$.

\textbf{Relative amplification.}
Since RL updates increase the probabilities of relatively better sampled trajectories \cite{li2026timerft}, when $A_k$ occurs, the update increases the probability of $\mathcal{S}$ more than that of $\mathcal{G}$:
\begin{equation}
    \frac{s_{k+1}}{s_k}
    \geq
    (1+\eta)\frac{p_{k+1}}{p_k},\eta>0.
    \label{eq:relative_amplification}
\end{equation}
We emphasize that these conditions characterize a possible failure mode and need not hold for every update of a specific RL method. Then, we can derive:

\begin{proposition}
\label{thm}
Under the above conditions, whenever $A_k$ occurs, after the update, the lower bound on the probability of the next low-quality-dominated update at time point $t$ increases:
\begin{equation}
    \label{eq_more}
    p(A_{k+1}\mid\mathcal{F}_{k+1})
    \geq
    \rho(1-h_{k+1})^M,
\end{equation}
where $h_k=\frac{p_k}{p_k+s_k}$. If such updates occur for $L$ consecutive iterations, then
\begin{equation}
    \label{eq_sc}
    p_{k+L}
    \leq
    \frac{1}{
    1+(1+\eta)^L s_k/p_k
    }
    \longrightarrow 0
    \qquad
    \text{as }L\rightarrow\infty.
\end{equation}
\end{proposition}
Proposition \ref{thm} characterizes a sufficient self-reinforcing mechanism for suboptimal collapse at a forecasting point. Intuitively, when the initial probability assigned to its ground-truth neighborhood is low (i.e., $p_0$ is low), a low-quality-dominated update ($A_0$) is more likely to occur. Once such an update occurs, Eq.\ref{eq_more} shows that the lower bound on the probability of the next low-quality-dominated update increases, making such updates more likely to persist. If this process persists, Eq.\ref{eq_sc} implies that the ground-truth-neighborhood probability at this forecasting point can approach zero, leading to suboptimal collapse.

\section{Empirical Analysis}
\label{Sec_analysis}

In this section, we provide empirical evidence consistent with the self-reinforcing process characterized in Proposition \ref{thm} and investigate the effect of adding the ground-truth trajectory to the sampled group during training.

\textbf{In the first experiment}, we empirically analyze suboptimal collapse. Specifically, we select three TSFMs as backbones—Moirai-small \cite{woo2024unified}, Toto \cite{cohen2024toto}, and Moirai-MoE$_s$ \cite{liu2024moirai}. And we apply TPO \cite{qi2025timehf}, TimeRFT \cite{li2026timerft}, and TS-GRPO, respectively, for RL post-training. TS-GRPO is a simple GRPO-based post-training method applicable to different TSFMs, with implementation details provided in Appendix "TS-GRPO". We first perform a complete RL post-training run. We then randomly select 1,000 training samples and identify the time points at which collapse does and does not occur in these samples. After training, a time point is considered collapsed if its ground-truth-neighborhood probability is below 0.001. The neighborhood radius is set to 1 for Loop Seattle and 0.1 for all other datasets. We then retrain the models and record, throughout training, the mean log probability assigned by the model’s output distribution to the ground-truth neighborhoods of these two groups of points. The results are shown in Fig.\ref{fig_motivation1}. We observe that: 1) regions of collapse typically have lower initial ground-truth-neighborhood probability, which further decreases during training; and 2) regions of non-collapse have higher initial probability, which continues to increase throughout training. These results are consistent with the conditional self-reinforcing process characterized in Proposition \ref{thm}: when the model initially struggles to sample high-quality trajectories in certain regions, a detrimental self-reinforcing cycle can emerge, eventually leading to suboptimal collapse.

\textbf{In the second experiment}, we examine the effect of adding the ground-truth trajectory to the sampled group during training. Specifically, the TSFMs, RL methods, and observed train data selection settings are the same as those described above. We train the models under two settings: with and without the ground-truth trajectory included. We clip the ground-truth reward at 1.5 times the maximum sampled reward within its group to prevent an excessively high ground-truth reward from compromising training stability. During training, we record three metrics: 1) the mean log probability assigned by the model’s output distribution to the ground-truth neighborhoods of these selected samples; 2) the sampling variance of the predictive trajectories for these selected samples; and 3) the forecasting performance of the trained model on the test set. The results are in Fig.\ref{fig_motivation2}, \ref{fig_motivation4} and \ref{fig_motivation3}. We observe that directly adding the ground-truth trajectory substantially increases the probability of the ground-truth neighborhood, thereby mitigating suboptimal collapse. However, both the sampling variance and test-set performance decrease. These results indicate that the ground-truth trajectory can serve as a reference for high-quality trajectories to mitigate suboptimal collapse, but should not directly participate in policy updates as a fixed high-reward sample.

\begin{figure*}[htbp]
    \centering
    \includegraphics[width=0.70\linewidth]{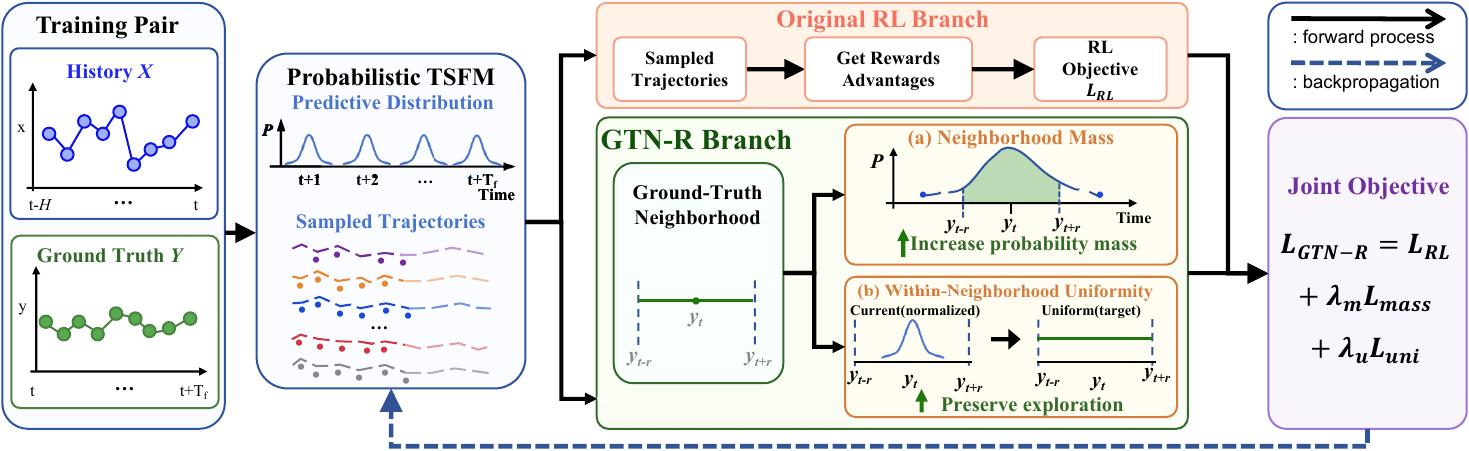}
    \caption{Overall framework of our method.}
    \label{method}
\end{figure*}

\section{Method}

In this section, we propose GTN-R, a regularization method for RL post-training of TSFMs. GTN-R uses the ground truth only to identify high-quality regions and increase the probability of sampling trajectories from them, without directly incorporating ground-truth trajectories into policy updates. This mitigates suboptimal collapse while preserving exploration, thereby improving performance. Specifically, we first introduce the overall pipeline, then describe the regularization constraints, and finally present the learning objective.

\subsection{Overall Pipeline}
\label{sec_pipe}

In this subsection, we introduce the overall pipeline of GTN-R. Fig.\ref{method} illustrates the overall training framework of GTN-R. Specifically, we adopt a TSFM capable of producing predictive distributions as the backbone model. Given a train date, we compute the Ground-Truth Neighborhood Regularization term, as detailed in the next subsection, and incorporate it into the original training objective. Finally, we update the model using the resulting objective under the RL paradigm.

\subsection{Ground-Truth Neighborhood Regularization}
\label{sec_regular}

In this subsection, we describe how to compute the proposed regularization terms based on the ground-truth neighborhood. Specifically, we first formally define the ground-truth neighborhood. We then introduce how to get the first regularization term: the neighborhood probability mass constraint. Finally, we introduce how to get the second regularization term: the within-neighborhood uniformity constraint.

\textbf{Firstly}, we formally define the ground-truth neighborhood. Given a historical series $X$ and the corresponding ground-truth future series:
\begin{equation}
    Y=\{y_1, y_2, ..., y_{T_f}\}.
\end{equation}
We construct a neighborhood centered at each ground-truth value $y_t$:
\begin{equation}
    \mathcal{N}_r(y_t) = [y_t - r,\, y_t + r],
\end{equation}
where $r$ is a hyperparameter. $\{\mathcal{N}_r(y_t)\}_{t=1}^{T_f}$ constitute the ground-truth neighborhood $\mathcal{N}_r(Y)$.

\textbf{Secondly}, we introduce how to compute the first regularization term, namely the neighborhood probability mass constraint. Specifically, let $p_{\theta,t}(z\mid\mathcal{C}_t^i)$ denote the probability density at $z$ of the predictive distribution at the $t$-th future time point of the $i$-th sampled trajectory. For one-shot forecasting, $\mathcal{C}_t^i=X$; for token-by-token forecasting, $\mathcal{C}_t^i$ additionally includes the previously generated segments. Next, we compute the probability mass assigned to the neighborhood $\mathcal{N}_r(y_t)$ under the context of the $i$-th trajectory:
\begin{equation}
    \label{eq_mass_i}
    p^i(\mathcal{N}_r(y_t)) = \int_{y_t-r}^{y_t+r} p_{\theta,t}(z \mid \mathcal{C}_t^i) \, dz.
\end{equation}
The integral in Eq.\ref{eq_mass_i} can be computed using Gauss–Legendre quadrature \cite{swarztrauber2003computing}. Based on this, we define the neighborhood probability mass constraint for the $i$-th sampled trajectory as the mean log-probability of $p^i(\mathcal{N}_r(y_t))$ over all future time points:
\begin{equation}
    \label{eq_sub_mass}
    \displaystyle
    \mathcal{L}_{\mathrm{mass}}^i
    = \frac{1}{T_f}\sum_{t=1}^{T_f}
    \log\!\left(p^i\!\left(\mathcal{N}_r(y_t)\right)\right)
\end{equation}
The reason for using the logarithm is that it imposes a stronger penalty on forecast points with extremely low probability, preventing them from being overlooked in the constraint. Our ablation study validates the rationality of this design.

For token-by-token forecasting, different sampled trajectories have different predictive distributions. We thus average the trajectory-wise constraints to get the final neighborhood probability mass constraint: $\mathcal{L}_{\mathrm{mass}}=\frac{1}{n}\sum_{i=1}^{n}\mathcal{L}_{\mathrm{mass}}^i, $ where $n$ is the number of trajectories sampled for each input during one update of the RL algorithm. For one-shot forecasting, the predictive distributions depend only on $X$ and are identical across sampled trajectories. That makes $\mathcal{L}_{\mathrm{mass}}^1=\cdots=\mathcal{L}_{\mathrm{mass}}^n$. Therefore, we simply set $\mathcal{L}_{\mathrm{mass}}=\mathcal{L}_{\mathrm{mass}}^i$ in this case. By maximizing $\mathcal{L}_{\mathrm{mass}}$, we encourage the model to assign more probability mass to the ground-truth neighborhoods under different trajectory contexts, thereby increasing the likelihood of sampling high-quality trajectories near the ground truth.

\textbf{Finally}, we describe how to compute the second regularization term, namely, the within-neighborhood uniformity constraint. Specifically, we first normalize the predictive density of the $i$-th sampled trajectory in $\mathcal{N}_r(y_t)$:
\begin{equation}
\resizebox{0.85\columnwidth}{!}{$
\widetilde{p}^i_{\theta,t}(z\mid\mathcal{C}_t^i)
=
p_{\theta,t}(z \mid \mathcal{C}_t^i)
/p^i(\mathcal{N}_r(y_t)),
\quad
z\in\mathcal{N}_r(y_t).
$}
\end{equation}

Meanwhile, we define the uniform distribution in $\mathcal{N}_r(y_t)$ as:
\begin{equation}
u_t(z)=1/2r,
\quad
z\in\mathcal{N}_r(y_t).
\end{equation}
Let $\widetilde{P}^i_{\theta,t}$ and $U_t$ denote the distributions corresponding to $\widetilde{p}^i_{\theta,t}(z\mid\mathcal{C}_t)$ and $u_t(z)$, respectively. Based on this, we define the within-neighborhood uniformity constraint for the $i$-th sampled trajectory as the negative mean KL divergence between these two distributions over all future time points:
\begin{equation}
\label{eq_dl}
    \mathcal{L}^i_{\text{uni}} = -\frac{1}{T_f} \sum_{t=1}^{T_f} D_{\text{KL}} \left( \widetilde{P}^i_{\theta,t}\parallel U_t \right).
\end{equation}
The $D_{\text{KL}}$ also can be computed using Gauss–Legendre quadrature \cite{swarztrauber2003computing}. Similar to $\mathcal{L}_{\mathrm{mass}}$, for token-by-token forecasting, $\mathcal{L}_{\mathrm{uni}}$ is defined as the mean of $\mathcal{L}_{\mathrm{uni}}^i$ over all sampled trajectories: $\mathcal{L}_{\mathrm{uni}}=\frac{1}{n}\sum_{i=1}^{n}\mathcal{L}_{\mathrm{uni}}^i$. For one-shot forecasting, $\mathcal{L}_{\mathrm{uni}}=\mathcal{L}_{\mathrm{uni}}^i$. By maximizing $\mathcal{L}_{\text{uni}}$, we encourage a uniform distribution within $\mathcal{N}_r(Y)$, preventing the probability mass from concentrating on only a few locations and thereby preserving the model’s exploration capability.

\begin{table*}[ht]
  \centering
  \resizebox{0.65\textwidth}{!}{
\begin{tabular}{l*{12}{c}}
\toprule
\multirow{1}[2]{*}{\centering Method}
& \multicolumn{2}{c}{\textbf{ETTm1}}
& \multicolumn{2}{c}{\textbf{ETTm2}}
& \multicolumn{2}{c}{\textbf{ETTh1}}
& \multicolumn{2}{c}{\textbf{ETTh2}}
& \multicolumn{2}{c}{\textbf{ECL}}
& \multicolumn{2}{c}{\textbf{Weather}} \\
\cmidrule(lr){2-3}  \cmidrule(lr){4-5}  \cmidrule(lr){6-7}  \cmidrule(lr){8-9}  \cmidrule(lr){10-11}  \cmidrule(lr){12-13}
& MSE & MAE & MSE & MAE & MSE & MAE & MSE & MAE & MSE & MAE & MSE & MAE\\
\midrule

\rowcolor{tabhighlight} {\moiraismall} & 0.448 & 0.409 & 0.300 & 0.341 & 0.416 & 0.427 & 0.354 & 0.381 & 0.233 & 0.320 & 0.268 & 0.279 \\
\hspace{0.5em} \textit{+ Full finetuning} & 0.367 & 0.382 & 0.273 & 0.316 & 0.415 & 0.428 & 0.352 & 0.378 & 0.193 & 0.279 & 0.228 & 0.254 \\
\hspace{0.5em} \textit{+ LoRA} & 0.370 & 0.383 & 0.272 & 0.314 & 0.414 & 0.427 & 0.353 & 0.380 & 0.192 & 0.279 & 0.225 & 0.252 \\
\hspace{0.5em} \textit{+ MSFT \citeyearpar{qiao2025multi}} & \underline{0.353} & 0.377 & \underline{0.250} & \underline{0.301} & 0.412 & 0.426 & 0.349 & 0.375 & \underline{0.187} & \underline{0.275} & \underline{0.215} & 0.248 \\
\hspace{0.5em} \textit{+ TPO \citeyearpar{qi2025timehf}} & 0.361 & 0.376 & 0.268 & 0.310 & 0.410 & 0.424 & 0.348 & 0.374 & 0.191 & 0.276 & 0.225 & 0.251 \\
\rowcolor{lightgreen} \hspace{0.5em} \textit{+ TPO \citeyearpar{qi2025timehf} + \textbf{GTN-R}} & 0.354 & \underline{0.373} & 0.258 & 0.303 & \underline{0.406} & \underline{0.421} & \underline{0.344} & \underline{0.371} & \textbf{0.179} & \textbf{0.267} & 0.216 & \underline{0.244} \\
\hspace{0.5em} \textit{+ TS-GRPO} & 0.358 & 0.375 & 0.265 & 0.308 & 0.416 & 0.429 & 0.351 & 0.377 & 0.200 & 0.287 & 0.226 & 0.252 \\
\rowcolor{lightgreen} \hspace{0.5em} \textit{+ TS-GRPO + \textbf{GTN-R}} & \textbf{0.345} & \textbf{0.369} & \textbf{0.244} & \textbf{0.293} & \textbf{0.403} & \textbf{0.419} & \textbf{0.339} & \textbf{0.367} & \textbf{0.179} & \textbf{0.267} & \textbf{0.210} & \textbf{0.239} \\
\midrule

\rowcolor{tabhighlight} {\moiraibase} & 0.382 & 0.388 & 0.281 & 0.326 & 0.412 & 0.424 & 0.356 & 0.388 & 0.188 & 0.274 & 0.246 & 0.265 \\
\hspace{0.5em} \textit{+ Full finetuning} & 0.368 & 0.371 & 0.258 & 0.307 & 0.408 & 0.424 & 0.357 & 0.385 & 0.173 & 0.264 & 0.232 & 0.258 \\
\hspace{0.5em} \textit{+ LoRA} & 0.361 & 0.370 & 0.259 & 0.307 & 0.408 & 0.423 & 0.356 & 0.387 & 0.172 & 0.263 & 0.230 & 0.260 \\
\hspace{0.5em} \textit{+ MSFT \citeyearpar{qiao2025multi}} & \textbf{0.332} & 0.369 & \underline{0.247} & 0.305 & 0.407 & 0.422 & 0.352 & 0.383 & 0.169 & 0.260 & \textbf{0.213} & \underline{0.244} \\
\hspace{0.5em} \textit{+ TPO \citeyearpar{qi2025timehf}} & 0.365 & 0.368 & 0.256 & 0.304 & \underline{0.402} & \underline{0.418} & 0.352 & 0.380 & 0.171 & 0.261 & 0.229 & 0.255 \\
\rowcolor{lightgreen} \hspace{0.5em} \textit{+ TPO \citeyearpar{qi2025timehf} + \textbf{GTN-R}} & 0.355 & \underline{0.359} & 0.248 & \underline{0.296} & \textbf{0.390} & \textbf{0.403} & \underline{0.345} & \underline{0.372} & \underline{0.164} & \underline{0.251} & 0.224 & 0.249 \\
\hspace{0.5em} \textit{+ TS-GRPO} & 0.357 & 0.364 & 0.253 & 0.301 & 0.415 & 0.429 & 0.349 & 0.378 & 0.179 & 0.267 & 0.222 & 0.248 \\
\rowcolor{lightgreen} \hspace{0.5em} \textit{+ TS-GRPO + \textbf{GTN-R}} & \underline{0.343} & \textbf{0.347} & \textbf{0.238} & \textbf{0.286} & \textbf{0.390} & \textbf{0.403} & \textbf{0.337} & \textbf{0.363} & \textbf{0.163} & \textbf{0.249} & \underline{0.217} & \textbf{0.243} \\
\midrule

\rowcolor{tabhighlight} {Toto} & 0.396 & 0.378 & 0.267 & 0.303 & 0.435 & 0.413 & 0.340 & 0.363 & 0.161 & 0.243 & 0.224 & 0.245 \\
\hspace{0.5em} \textit{+ Full finetuning} & 0.376 & 0.370 & 0.258 & 0.301 & 0.428 & 0.409 & 0.339 & 0.361 & 0.159 & 0.240 & 0.222 & 0.243 \\
\hspace{0.5em} \textit{+ LoRA} & 0.378 & 0.374 & 0.260 & 0.301 & 0.426 & 0.407 & 0.338 & 0.360 & 0.158 & 0.239 & 0.221 & 0.242 \\
\hspace{0.5em} \textit{+ MSFT \citeyearpar{qiao2025multi}} & \underline{0.357} & 0.367 & 0.249 & 0.296 & 0.422 & 0.406 & 0.337 & 0.358 & \underline{0.155} & \underline{0.236} & 0.217 & 0.240 \\
\hspace{0.5em} \textit{+ TS-GRPO} & 0.358 & \underline{0.362} & \underline{0.248} & \underline{0.293} & \underline{0.418} & \underline{0.403} & \underline{0.334} & \underline{0.355} & \underline{0.155} & \underline{0.236} & \underline{0.216} & \underline{0.239} \\
\rowcolor{lightgreen} \hspace{0.5em} \textit{+ TS-GRPO + \textbf{GTN-R}} & \textbf{0.343} & \textbf{0.349} & \textbf{0.237} & \textbf{0.284} & \textbf{0.409} & \textbf{0.399} & \textbf{0.329} & \textbf{0.351} & \textbf{0.152} & \textbf{0.233} & \textbf{0.210} & \textbf{0.235} \\
\midrule

\rowcolor{tabhighlight} {UniTS} & 0.713 & 0.552 & 0.321 & 0.355 & 0.527 & 0.491 & 0.405 & 0.417 & 0.432 & 0.488 & 0.291 & 0.313 \\
\hspace{0.5em} \textit{+ Full finetuning} & 0.395 & 0.405 & 0.296 & 0.338 & 0.442 & 0.435 & 0.386 & 0.408 & 0.214 & 0.283 & 0.257 & 0.282 \\
\hspace{0.5em} \textit{+ LoRA} & 0.393 & 0.405 & 0.296 & 0.338 & 0.437 & 0.434 & 0.384 & 0.407 & 0.188 & 0.281 & 0.250 & 0.278 \\
\hspace{0.5em} \textit{+ MSFT \citeyearpar{qiao2025multi}} & 0.390 & 0.403 & 0.286 & 0.333 & 0.434 & 0.430 & 0.380 & 0.405 & 0.184 & 0.279 & 0.241 & 0.272 \\
\hspace{0.5em} \textit{+ TPO \citeyearpar{qi2025timehf}} & 0.386 & 0.399 & 0.283 & 0.329 & 0.429 & 0.426 & 0.376 & 0.401 & 0.182 & 0.276 & 0.239 & 0.270 \\
\rowcolor{lightgreen} \hspace{0.5em} \textit{+ TPO \citeyearpar{qi2025timehf} + \textbf{GTN-R}} & \underline{0.381} & \underline{0.396} & 0.280 & \underline{0.326} & \underline{0.423} & \underline{0.421} & \underline{0.370} & \underline{0.397} & \underline{0.180} & \underline{0.274} & \underline{0.236} & \underline{0.267} \\
\hspace{0.5em} \textit{+ TS-GRPO} & 0.386 & 0.399 & \underline{0.279} & 0.327 & 0.431 & 0.428 & 0.373 & 0.398 & \underline{0.180} & \underline{0.274} & 0.238 & 0.269 \\
\rowcolor{lightgreen} \hspace{0.5em} \textit{+ TS-GRPO + \textbf{GTN-R}} & \textbf{0.374} & \textbf{0.391} & \textbf{0.275} & \textbf{0.323} & \textbf{0.414} & \textbf{0.416} & \textbf{0.364} & \textbf{0.393} & \textbf{0.177} & \textbf{0.271} & \textbf{0.232} & \textbf{0.264} \\
\bottomrule
\end{tabular}
}
\caption{Forecasting results averaged from four prediction lengths $\in$ \{96, 192, 336, 720\}. Most baseline results are from \cite{qiao2025multi}; the rest are obtained by running the source code on the datasets.}
\label{tab:lsf_avg_mse_mae}
\end{table*}

\begin{table}[t]
  \centering
  \resizebox{\columnwidth}{!}{
  \begin{tabular}{lcccccccc}
    \toprule
    Method & \multicolumn{2}{c}{5\%} & \multicolumn{2}{c}{20\%} & \multicolumn{2}{c}{50\%} & \multicolumn{2}{c}{100\%} \\
    \cmidrule(lr){2-3} \cmidrule(lr){4-5} \cmidrule(lr){6-7} \cmidrule(lr){8-9}
    \multicolumn{9}{c}{\textbf{Loop Seattle}} \\
    & \shortstack{MSE\\($\times 10^{1}$)} & \shortstack{MAE\\($\times 10^{0}$)} & \shortstack{MSE\\($\times 10^{1}$)} & \shortstack{MAE\\($\times 10^{0}$)} & \shortstack{MSE\\($\times 10^{1}$)} & \shortstack{MAE\\($\times 10^{0}$)} & \shortstack{MSE\\($\times 10^{1}$)} & \shortstack{MAE\\($\times 10^{0}$)} \\
    \midrule
    \rowcolor{tabhighlight} $MOIRAI-MoE_s$ & 4.186 & 3.971 & 4.186 & 3.971 & 4.186 & 3.971 & 4.186 & 3.971 \\
    + TimeSFT & 2.996 & 3.625 & 2.614 & 3.553 & 2.535 & 3.379 & 2.421 & 3.448 \\
    + TimeLoRA & 3.011 & 3.622 & 2.697 & 3.584 & 2.565 & 3.370 & 2.434 & 3.464 \\
    + TS-GRPO & 2.911 & 3.601 & 2.424 & 3.428 & 2.314 & 3.302 & 2.254 & 3.265 \\
    \rowcolor{lightgreen} + TS-GRPO + \textbf{GTN-R} & 2.852 & 3.553 & 2.354 & 3.366 & 2.248 & 3.251 & 2.213 & 3.224 \\
    + TimeRFT & \underline{2.757} & \underline{3.400} & \underline{2.193} & \underline{3.091} & \underline{2.082} & \underline{3.019} & \underline{2.032} & \underline{3.021} \\
    \rowcolor{lightgreen} + TimeRFT + \textbf{GTN-R} & \textbf{2.702} & \textbf{3.357} & \textbf{2.125} & \textbf{3.012} & \textbf{2.013} & \textbf{2.956} & \textbf{1.994} & \textbf{2.978} \\
    \midrule
    \multicolumn{9}{c}{\textbf{ENTSO-e Load}} \\
    & \shortstack{MSE\\($\times 10^{5}$)} & \shortstack{MAE\\($\times 10^{2}$)} & \shortstack{MSE\\($\times 10^{5}$)} & \shortstack{MAE\\($\times 10^{2}$)} & \shortstack{MSE\\($\times 10^{5}$)} & \shortstack{MAE\\($\times 10^{2}$)} & \shortstack{MSE\\($\times 10^{5}$)} & \shortstack{MAE\\($\times 10^{2}$)} \\
    \midrule
    \rowcolor{tabhighlight} $MOIRAI-MoE_s$ & 22.832 & 11.876 & 22.832 & 11.876 & 22.832 & 11.876 & 22.832 & 11.876 \\
    + TimeSFT & 16.066 & 8.860 & 6.465 & 5.561 & 6.682 & 5.800 & 5.379 & 5.232 \\
    + TimeLoRA & 15.801 & 8.836 & 6.291 & 5.498 & 6.770 & 5.869 & 5.348 & 5.209 \\
    + TS-GRPO & 16.146 & 8.941 & 6.520 & 5.621 & 4.312 & 4.619 & 4.095 & 4.532 \\
    \rowcolor{lightgreen} + TS-GRPO + \textbf{GTN-R} & 16.013 & 8.852 & 6.452 & 5.560 & 4.237 & 4.571 & 3.999 & 4.412 \\
    + TimeRFT & \underline{15.329} & \underline{8.643} & \underline{4.196} & \underline{4.525} & \underline{3.867} & \underline{4.189} & \underline{3.767} & \underline{4.230} \\
    \rowcolor{lightgreen} + TimeRFT + \textbf{GTN-R} & \textbf{15.156} & \textbf{8.593} & \textbf{4.132} & \textbf{4.469} & \textbf{3.734} & \textbf{4.115} & \textbf{3.710} & \textbf{4.187} \\
    \bottomrule
  \end{tabular}
  }
  \caption{More Forecasting results. The baseline results derive from TimeRFT \cite{li2026timerft}. The percentages in the table indicate the proportion of the training data used. The experimental setup follows Table 2 in TimeRFT.}
  \label{tab:timerft_full_ratio_results}
\end{table}

\begin{table}[ht]
  \centering
  \resizebox{\columnwidth}{!}{
\begin{tabular}{l*{8}{c}}
\toprule
\multirow{1}[2]{*}{\centering Method}
& \multicolumn{2}{c}{\textbf{ETTm1$\rightarrow$ETTm2}}
& \multicolumn{2}{c}{\textbf{ETTm2$\rightarrow$ETTm1}}
& \multicolumn{2}{c}{\textbf{ETTm2$\rightarrow$ETTh2}}
& \multicolumn{2}{c}{\textbf{ETTh1$\rightarrow$Weather}} \\
\cmidrule(lr){2-3}  \cmidrule(lr){4-5}  \cmidrule(lr){6-7}  \cmidrule(lr){8-9}
& MSE & MAE & MSE & MAE & MSE & MAE & MSE & MAE \\
\midrule

\rowcolor{tabhighlight} {\moiraismall} & 0.300 & 0.341 & 0.448 & 0.409 & 0.354 & 0.381 & 0.268 & 0.279 \\
\hspace{0.5em} \textit{+ Full finetuning} & 0.293 & 0.339 & 0.453 & 0.412 & 0.358 & 0.384 & 0.263 & 0.276 \\
\hspace{0.5em} \textit{+ LoRA} & 0.321 & 0.396 & 0.452 & 0.413 & 0.359 & 0.384 & 0.264 & 0.277 \\
\hspace{0.5em} \textit{+ MSFT \citeyearpar{qiao2025multi}} & 0.288 & 0.336 & 0.469 & 0.419 & 0.355 & 0.381 & 0.259 & 0.272 \\
\hspace{0.5em} \textit{+ TPO \citeyearpar{qi2025timehf}} & 0.290 & 0.335 & 0.443 & 0.407 & 0.354 & 0.380 & 0.261 & 0.274 \\
\rowcolor{lightgreen} \hspace{0.5em} \textit{+ TPO \citeyearpar{qi2025timehf} + \textbf{GTN-R}} & \textbf{0.278} & \textbf{0.332} & \textbf{0.437} & \textbf{0.403} & \underline{0.349} & \underline{0.376} & \underline{0.257} & \underline{0.270} \\
\hspace{0.5em} \textit{+ TS-GRPO} & 0.289 & 0.335 & 0.446 & 0.407 & 0.351 & 0.381 & 0.259 & 0.273 \\
\rowcolor{lightgreen} \hspace{0.5em} \textit{+ TS-GRPO + \textbf{GTN-R}} & \underline{0.280} & \underline{0.334} & \underline{0.438} & \underline{0.404} & \textbf{0.346} & \textbf{0.374} & \textbf{0.256} & \textbf{0.269} \\
\midrule

\rowcolor{tabhighlight} {\moiraibase} & 0.281 & 0.326 & 0.382 & 0.388 & 0.356 & 0.388 & 0.246 & 0.265 \\
\hspace{0.5em} \textit{+ Full finetuning} & 0.274 & 0.324 & 0.387 & 0.391 & 0.360 & 0.391 & 0.242 & 0.262 \\
\hspace{0.5em} \textit{+ LoRA} & 0.300 & 0.377 & 0.384 & 0.389 & 0.360 & 0.393 & 0.242 & 0.262 \\
\hspace{0.5em} \textit{+ MSFT \citeyearpar{qiao2025multi}} & 0.269 & 0.320 & 0.400 & 0.397 & 0.358 & 0.390 & 0.238 & 0.258 \\
\hspace{0.5em} \textit{+ TPO \citeyearpar{qi2025timehf}} & 0.271 & 0.324 & 0.378 & 0.385 & 0.356 & 0.386 & 0.238 & 0.258 \\
\rowcolor{lightgreen} \hspace{0.5em} \textit{+ TPO \citeyearpar{qi2025timehf} + \textbf{GTN-R}} & \underline{0.265} & \underline{0.315} & \textbf{0.371} & \textbf{0.380} & \underline{0.351} & \underline{0.380} & \textbf{0.234} & \textbf{0.254} \\
\hspace{0.5em} \textit{+ TS-GRPO} & 0.270 & 0.322 & 0.378 & 0.386 & 0.354 & 0.386 & 0.239 & 0.259 \\
\rowcolor{lightgreen} \hspace{0.5em} \textit{+ TS-GRPO + \textbf{GTN-R}} & \textbf{0.263} & \textbf{0.314} & \underline{0.373} & \underline{0.382} & \textbf{0.345} & \textbf{0.375} & \underline{0.236} & \underline{0.256} \\
\bottomrule
\end{tabular}
}
\caption{Zero-shot results averaged over prediction lengths ${96,192,336,720}$. }
  \label{tab:zero_shot_avg_mse_mae}
\end{table}

\begin{figure*}[htbp]
    \centering
    \subfloat[]{\includegraphics[width=0.33\textwidth]{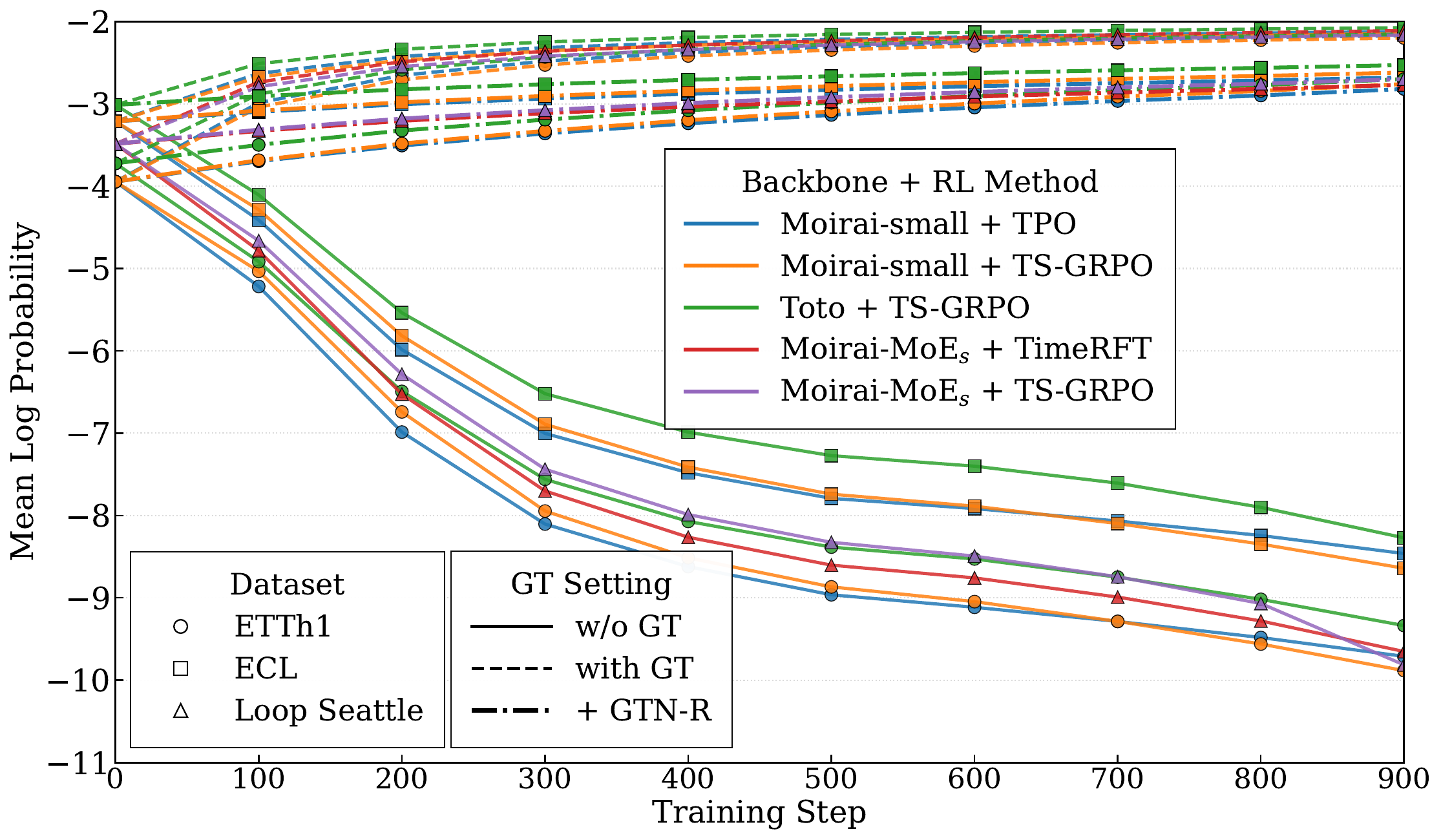}
    \label{fig_ana1}}
    \hfil
    \subfloat[]{\includegraphics[width=0.33\textwidth]{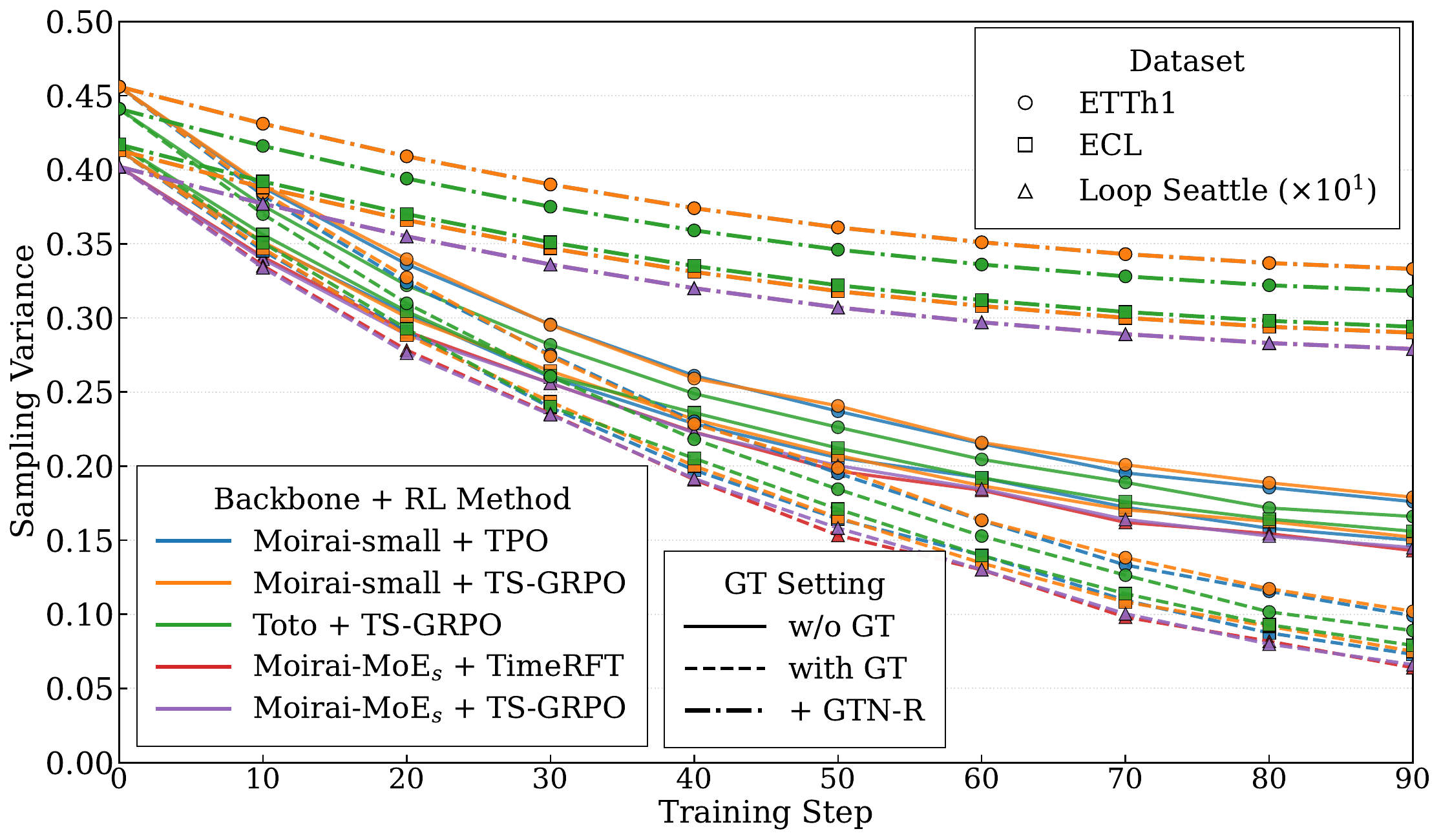}
    \label{fig_ana2}}
    \hfil
    \subfloat[]{\includegraphics[width=0.33\textwidth]{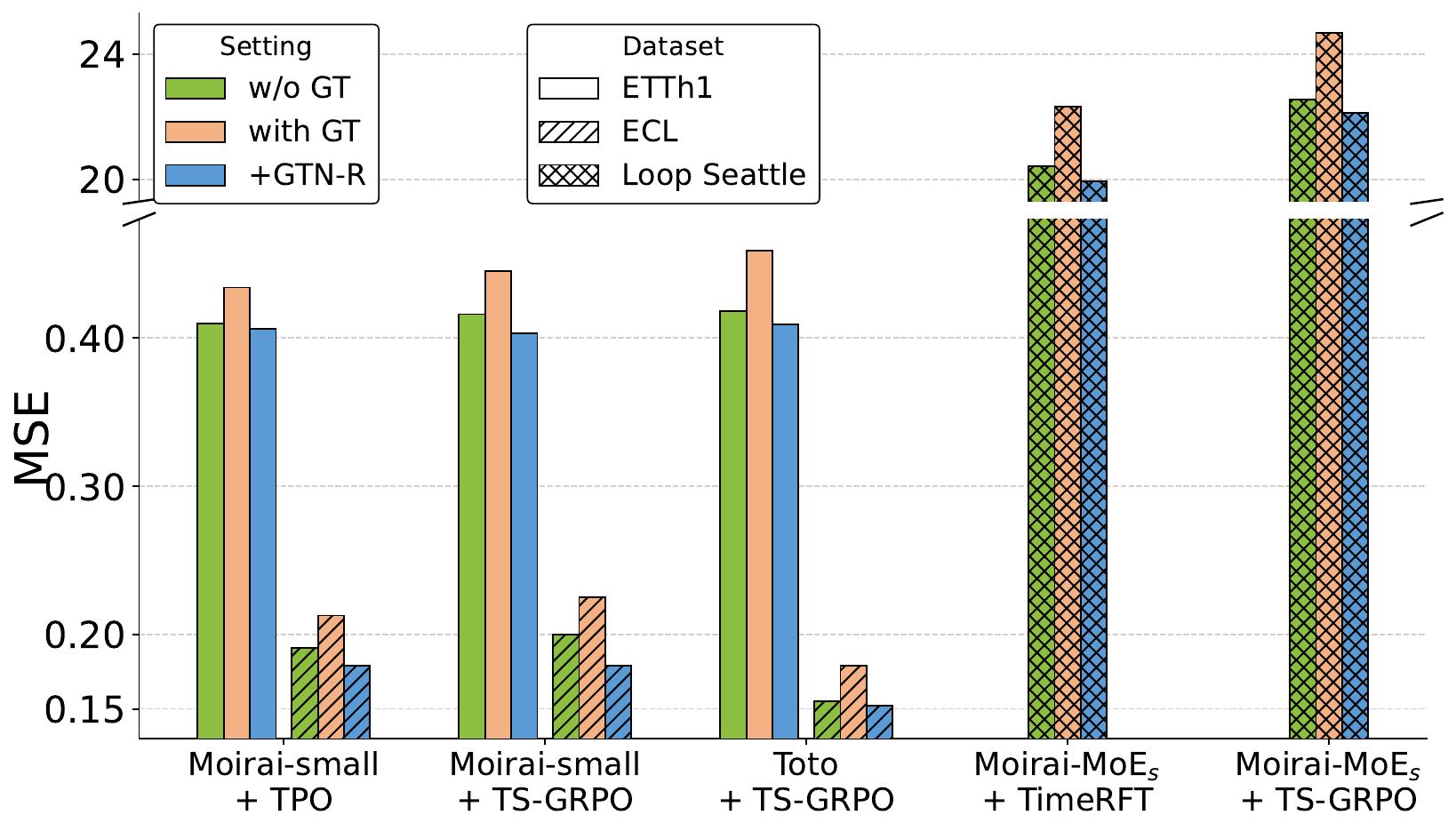}
    \label{fig_ana3}}
    \hfil
    \caption{(a) Mean ground-truth-neighborhood log probability during training. (b) and sampling variance during training. (c) Test-set forecasting performance. In (b), $\times 10^1$ indicates that the actual variances of the dataset are 10 times the plotted values.}
    \label{fig_ana}
\end{figure*}

\begin{figure}[t]
    \centering
    \includegraphics[width=\columnwidth]{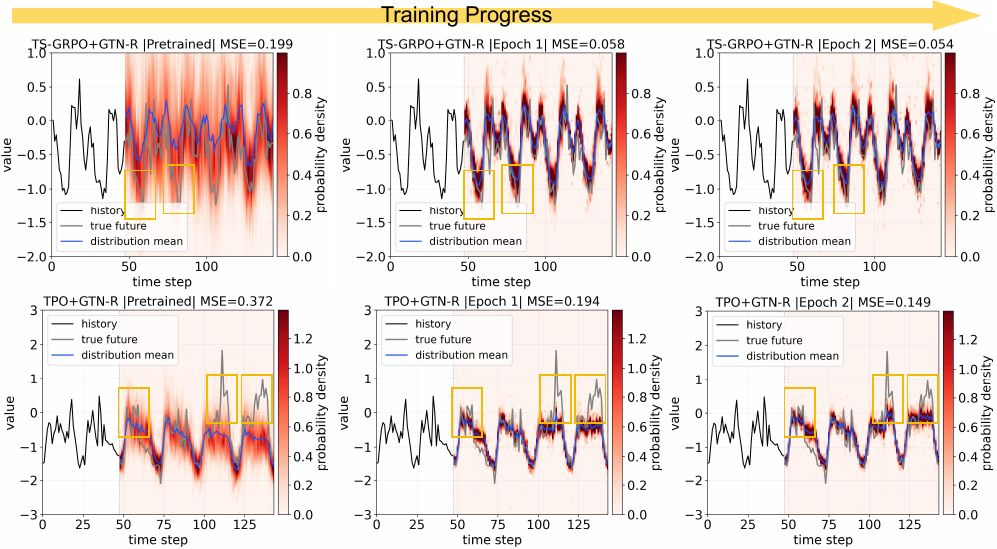}
    \caption{The output distributions for the training samples in Fig.\ref{collapse} after incorporating GTN-R. More visualization results are provided in Appendix "More Visualization".
    }
    \label{collapse_s}
\end{figure}

\subsection{Learning Objective of GTN-R}
\label{sec_objective}

After obtaining the neighborhood probability mass constraint and the within-neighborhood uniformity constraint, we incorporate both into the original RL objective, yielding the GTN-R learning objective to be maximized:
\begin{equation}
\mathcal{L}_{GTN-R}
=
\mathcal{L}_{RL}
+
\lambda_m\mathcal{L}_{mass}
+
\lambda_u\mathcal{L}_{uni},
\end{equation}
where, $\mathcal{L}_{\text{RL}}$ denotes the original RL objective, while $\lambda_m$ and $\lambda_u$ are the weights of the two regularization terms. Maximizing this objective enables the model to mitigate suboptimal collapse while preserving its autonomous exploration capability, thereby further improving forecasting performance.

\section{Experiments}

In this section, we empirically evaluate GTN-R. We first introduce the experimental setup, then present comparative results of multiple datasets and visualize how GTN-R affects the model’s output distributions during training. Additional visualizations, ablations, computational-cost, and hyperparameter-sensitivity are provided in the Appendix.

\subsection{Experimental Settings}
\label{Sec_set}
\textbf{Datasets: }

For general time series forecasting, we use eight real-world datasets: ETT (ETTh1, ETTh2, ETTm1, and ETTm2), ECL, Weather, Loop Seattle, and ENTSO-e Load. For zero-shot forecasting, we use the ETT and Weather datasets. The data splits for Loop Seattle and ENTSO-e Load follow TimeRFT \cite{li2026timerft}, while those for the remaining datasets follow \cite{qiao2025multi}.

\textbf{Baselines: }

We mainly follow the post-training evaluation framework of \cite{qiao2025multi} and adopt Moirai \cite{woo2024unified}, Toto \cite{cohen2024toto}, Moirai-MoE \cite{liu2024moirai}, and UniTS \cite{gao2024units} as the time series foundation models. Following \cite{qiao2025multi}, we use the x128 pretrained checkpoint for UniTS. Since UniTS supports only deterministic forecasting, when integrated with RL methods, we modify and retrain its output layer to produce predictive distributions; further details are in Appendix "Probabilistic Adaptation of UniTS". The compared post-training methods include full-parameter supervised fine-tuning (Full SFT), LoRA, MSFT \cite{qiao2025multi}, TPO \cite{qi2025timehf}, TimeRFT \cite{li2026timerft}, and our proposed TS-GRPO. Implementation details of TS-GRPO are provided in Appendix "TS-GRPO". TPO is applicable only to TSFMs that adopt the one-shot forecasting paradigm, whereas TimeRFT is specifically designed for Moirai-MoE. Accordingly, we apply TPO only to Moirai and UniTS, and TimeRFT only to Moirai-MoE. In contrast, TS-GRPO can be applied to any TSFM with probabilistic outputs. Moreover, because TimeRFT involves a complex training pipeline and numerous hyperparameters, we compare against it only on the Loop Seattle and ENTSO-e Load datasets, for which its training scripts are publicly available. We do not compare with conventional TSF methods, such as iTransformer \cite{liu2023itransformer}, PatchTST \cite{PatchTST}, DLinear \cite{DLinear}, TimeMixer \cite{wang2024timemixer}, and SimpleTM \cite{chen2025simpletm}, because pretrained TSFMs already achieve comparable performance, making comparisons with post-trained TSFMs unfair.

\textbf{Implementation Details:}

All experiments are implemented in PyTorch \cite{imambi2021pytorch} on an NVIDIA RTX 4090 GPU. We use the Adam optimizer \cite{kingma2014adam}. The hyperparameters $\lambda_m$ and $\lambda_u$ are both set to 1. The neighborhood radius $r$ is set to 1 for Loop Seattle, 100 for ENTSO-e Load, and 0.1 for all other datasets. These values are selected based on the hyperparameter sensitivity results. Other settings, including the learning rate and batch size, follow the best configurations of the corresponding post-training methods. The maximum number of training epochs is set to 40. Early stopping is triggered when the validation loss does not decrease for three consecutive evaluation rounds.

\subsection{Forecasting Performance}
\label{Sec_eva}

We evaluate GTN-R under both standard and zero-shot forecasting settings. In the standard setting, we compare our method with other methods, with results reported in Tables \ref{tab:lsf_avg_mse_mae} and \ref{tab:timerft_full_ratio_results}. In the zero-shot setting, the model is trained on one dataset and evaluated on another, with results shown in Table \ref{tab:zero_shot_avg_mse_mae}. All results are averaged over five runs with different random seeds. For each base method, the best post-training result is highlighted in bold, while the second-best is underlined. Overall, incorporating GTN-R can improve the performance of existing RL post-training methods.

\subsection{Method Analysis}
\label{Sec_m_ana}

In this subsection, we further validate the effectiveness of the proposed method by visualizing how the model’s output distribution evolves during training after incorporating GTN-R. The experimental setup and the selection of samples for visualization are consistent with those in Fig.\ref{fig_motivation2}. During training, we record the following three metrics: 1) the mean log probability assigned by the model’s output distribution to the ground-truth neighborhoods of the selected training samples; 2) the sampling variance of the predictive trajectories for these samples; and 3) the forecasting performance of the trained model on the test set. We compare three settings: standard training, directly adding the ground-truth trajectory to the sampled group, and training with GTN-R. The results are presented in Figs.\ref{fig_ana}. As shown in Fig.\ref{fig_ana1}, GTN-R prevents the ground-truth-neighborhood probability mass from decreasing, thereby mitigating suboptimal collapse. Fig.\ref{fig_ana2} further shows that its sampling variance declines much less than in the other two settings, indicating better preservation of the exploration ability. Consequently, GTN-R achieves superior test-set forecasting performance, as shown in Fig.\ref{fig_ana3}.

Furthermore, we visualize the evolution of the predictive distributions for the training samples in Fig.\ref{collapse} after incorporating GTN-R. We use the same random seed in the experiments to ensure consistent training dynamics. The results are in Fig.~\ref{collapse_s}. Comparing Figs.\ref{collapse} and \ref{collapse_s}, we observe that GTN-R mitigates suboptimal collapse as training progresses.

\section{Conclusion}

This paper identifies suboptimal collapse in RL post-training of TSFMs and analyzes its self-reinforcing mechanism. To mitigate its negative impact on training, we propose GTN-R, which increases the probability mass assigned to the ground-truth neighborhood while maintaining a uniform distribution within it. This alleviates suboptimal collapse while preserving the model’s exploration capability. Experiments across multiple models, datasets, and RL methods demonstrate that GTN-R consistently improves forecasting performance.

\clearpage
\bibliography{aaai2027}

@article{li2026timerft,
  title={TimeRFT: Stimulating Generalizable Time Series Forecasting for TSFMs via Reinforcement Finetuning},
  author={Li, Siyang and Chen, Yize and Zhu, Zijie and Pan, Yuxin and Guo, Yan and Huang, Ming and Xiong, Hui},
  journal={arXiv preprint arXiv:2605.00015},
  year={2026}
}

@article{PatchTST,
  title={A Time Series is Worth 64 Words: Long-term Forecasting with Transformers},
  author={Nie, Yuqi and Nguyen, Nam H and Sinthong, Phanwadee and Kalagnanam, Jayant},
  journal={ICLR},
  year={2023}
}

@article{kingma2014adam,
  title={Adam: A method for stochastic optimization},
  author={Kingma, Diederik P and Ba, Jimmy},
  journal={arXiv preprint arXiv:1412.6980},
  year={2014}
}

@article{yan2026s,
  title={S-GRPO: Unified Post-Training for Large Vision-Language Models},
  author={Yan, Yuming and Tang, Kai and Chen, Sihong and Xu, Ke and Hu, Dan and Yu, Qun and Hu, Pengfei},
  journal={arXiv preprint arXiv:2604.16557},
  year={2026}
}

@article{liu2024timer,
  title={Timer: Generative pre-trained transformers are large time series models},
  author={Liu, Yong and Zhang, Haoran and Li, Chenyu and Huang, Xiangdong and Wang, Jianmin and Long, Mingsheng},
  journal={arXiv preprint arXiv:2402.02368},
  year={2024}
}

@article{cohen2024toto,
  title={Toto: Time series optimized transformer for observability},
  author={Cohen, Ben and Khwaja, Emaad and Wang, Kan and Masson, Charles and Ram{\'e}, Elise and Doubli, Youssef and Abou-Amal, Othmane},
  journal={arXiv preprint arXiv:2407.07874},
  year={2024}
}

@article{liu2024moirai,
  title={Moirai-moe: Empowering time series foundation models with sparse mixture of experts},
  author={Liu, Xu and Liu, Juncheng and Woo, Gerald and Aksu, Taha and Liang, Yuxuan and Zimmermann, Roger and Liu, Chenghao and Savarese, Silvio and Xiong, Caiming and Sahoo, Doyen},
  journal={arXiv preprint arXiv:2410.10469},
  year={2024}
}

@article{shao2024deepseekmath,
  title={Deepseekmath: Pushing the limits of mathematical reasoning in open language models},
  author={Shao, Zhihong and Wang, Peiyi and Zhu, Qihao and Xu, Runxin and Song, Junxiao and Bi, Xiao and Zhang, Haowei and Zhang, Mingchuan and Li, YK and Wu, Yang and others},
  journal={arXiv preprint arXiv:2402.03300},
  year={2024}
}

@article{ouyang2022training,
  title={Training language models to follow instructions with human feedback},
  author={Ouyang, Long and Wu, Jeffrey and Jiang, Xu and Almeida, Diogo and Wainwright, Carroll and Mishkin, Pamela and Zhang, Chong and Agarwal, Sandhini and Slama, Katarina and Ray, Alex and others},
  journal={Advances in neural information processing systems},
  volume={35},
  pages={27730--27744},
  year={2022}
}

@article{qi2025timehf,
  title={Timehf: Billion-scale time series models guided by human feedback},
  author={Qi, Yongzhi and Hu, Hao and Lei, Dazhou and Zhang, Jianshen and Shi, Zhengxin and Huang, Yulin and Chen, Zhengyu and Lin, Xiaoming and Shen, Zuo-Jun Max},
  journal={arXiv preprint arXiv:2501.15942},
  year={2025}
}

@inproceedings{das2024decoder,
  title={A decoder-only foundation model for time-series forecasting},
  author={Das, Abhimanyu and Kong, Weihao and Sen, Rajat and Zhou, Yichen},
  booktitle={Forty-first International Conference on Machine Learning},
  year={2024}
}

@article{gao2024units,
  title={Units: A unified multi-task time series model},
  author={Gao, Shanghua and Koker, Teddy and Queen, Owen and Hartvigsen, Tom and Tsiligkaridis, Theodoros and Zitnik, Marinka},
  journal={Advances in Neural Information Processing Systems},
  volume={37},
  pages={140589--140631},
  year={2024}
}

@article{goswami2024moment,
  title={Moment: A family of open time-series foundation models},
  author={Goswami, Mononito and Szafer, Konrad and Choudhry, Arjun and Cai, Yifu and Li, Shuo and Dubrawski, Artur},
  journal={arXiv preprint arXiv:2402.03885},
  year={2024}
}

@article{swarztrauber2003computing,
  title={On computing the points and weights for Gauss--Legendre quadrature},
  author={Swarztrauber, Paul N},
  journal={SIAM Journal on Scientific Computing},
  volume={24},
  number={3},
  pages={945--954},
  year={2003},
  publisher={SIAM}
}

@article{qiao2025multi,
  title={Multi-Scale Finetuning for Encoder-based Time Series Foundation Models},
  author={Qiao, Zhongzheng and Liu, Chenghao and Zhang, Yiming and Jin, Ming and Pham, Quang and Wen, Qingsong and Suganthan, PN and Jiang, Xudong and Ramasamy, Savitha},
  journal={arXiv preprint arXiv:2506.14087},
  year={2025}
}

@String{Computing = "Computing" }

@String{Springer = "Springer-Verlag" }

@ArtifactSoftware{R,
    title = {R: A Language and Environment for Statistical Computing},
    author = {{R Core Team}},
    organization = {R Foundation for Statistical Computing},
    address = {Vienna, Austria},
    year = {2019},
    url = {https://www.R-project.org/},
}

@article{DLinear,
  title={Are Transformers Effective for Time Series Forecasting?},
  author={Ailing Zeng and Muxi Chen and Lei Zhang and Qiang Xu},
  journal={AAAI},
  year={2023}
}

@article{wang2024timemixer,
  title={Timemixer: Decomposable multiscale mixing for time series forecasting},
  author={Wang, Shiyu and Wu, Haixu and Shi, Xiaoming and Hu, Tengge and Luo, Huakun and Ma, Lintao and Zhang, James Y and Zhou, Jun},
  journal={arXiv preprint arXiv:2405.14616},
  year={2024}
}

@inproceedings{chen2025simpletm,
  title={SimpleTM: A simple baseline for multivariate time series forecasting},
  author={Chen, Hui and Luong, Viet and Mukherjee, Lopamudra and Singh, Vikas},
  booktitle={The Thirteenth International Conference on Learning Representations},
  year={2025}
}

@article{liu2023itransformer,
  title={itransformer: Inverted transformers are effective for time series forecasting},
  author={Liu, Yong and Hu, Tengge and Zhang, Haoran and Wu, Haixu and Wang, Shiyu and Ma, Lintao and Long, Mingsheng},
  journal={arXiv preprint arXiv:2310.06625},
  year={2023}
}

@article{abhishek2012weather,
  title={Weather forecasting model using artificial neural network},
  author={Abhishek, Kumar and Singh, MP and Ghosh, Saswata and Anand, Abhishek},
  journal={Procedia Technology},
  volume={4},
  pages={311--318},
  year={2012},
  publisher={Elsevier}
}

@article{miller2024survey,
  title={A survey of deep learning and foundation models for time series forecasting},
  author={Miller, John A and Aldosari, Mohammed and Saeed, Farah and Barna, Nasid Habib and Rana, Subas and Arpinar, I Budak and Liu, Ninghao},
  journal={arXiv preprint arXiv:2401.13912},
  year={2024}
}

@article{li2022dmgan,
  title={DMGAN: Dynamic multi-hop graph attention network for traffic forecasting},
  author={Li, Rui and Zhang, Fan and Li, Tong and Zhang, Ning and Zhang, Tingting},
  journal={IEEE Transactions on Knowledge and Data Engineering},
  year={2022},
  publisher={IEEE}
}

@article{fang2023stwave+,
  title={STWave+: A Multi-Scale Efficient Spectral Graph Attention Network With Long-Term Trends for Disentangled Traffic Flow Forecasting},
  author={Fang, Yuchen and Qin, Yanjun and Luo, Haiyong and Zhao, Fang and Zheng, Kai},
  journal={IEEE Transactions on Knowledge and Data Engineering},
  year={2023},
  publisher={IEEE}
}

@article{hamilton2007ski,
  title={Ski areas, weather and climate: time series models for New England case studies},
  author={Hamilton, Lawrence C and Brown, Cliff and Keim, Barry D},
  journal={International Journal of Climatology},
  year={2007},
  publisher={Wiley}
}

@article{shekhar2007adaptive,
  title={Adaptive seasonal time series models for forecasting short-term traffic flow},
  author={Shekhar, Shashank and Williams, Billy M},
  journal={Transportation Research Record},
  volume={2024},
  number={1},
  pages={116--125},
  year={2007},
  publisher={SAGE Publications Sage CA: Los Angeles, CA}
}

@article{deb2017review,
  title={A review on time series forecasting techniques for building energy consumption},
  author={Deb, Chirag and Zhang, Fan and Yang, Junjing and Lee, Siew Eang and Shah, Kwok Wei},
  journal={Renewable and Sustainable Energy Reviews},
  volume={74},
  pages={902--924},
  year={2017},
  publisher={Elsevier}
}

@article{wang2025benefits,
  title={Benefits and pitfalls of reinforcement learning for language model planning: a theoretical perspective},
  author={Wang, Siwei and Shen, Yifei and Sun, Haoran and Feng, Shi and Teng, Shang-Hua and Dong, Li and Hao, Yaru and Chen, Wei},
  journal={arXiv preprint arXiv:2509.22613},
  year={2025}
}

@article{weltevrede2024explore,
  title={Explore-Go: Leveraging Exploration for Generalisation in Deep Reinforcement Learning},
  author={Weltevrede, Max and Kaubek, Felix and Spaan, Matthijs TJ and B{\"o}hmer, Wendelin},
  journal={arXiv preprint arXiv:2406.08069},
  year={2024}
}

@article{wang2022st,
  title={St-expertnet: A deep expert framework for traffic prediction},
  author={Wang, Hongjun and Chen, Jiyuan and Fan, Zipei and Zhang, Zhiwen and Cai, Zekun and Song, Xuan},
  journal={IEEE Transactions on Knowledge and Data Engineering},
  year={2022},
  publisher={IEEE}
}

@article{campbell2005weather,
  title={Weather forecasting for weather derivatives},
  author={Campbell, Sean D and Diebold, Francis X},
  journal={Journal of the American Statistical Association},
  volume={100},
  number={469},
  pages={6--16},
  year={2005},
  publisher={Taylor \& Francis}
}

@article{karevan2020transductive,
  title={Transductive LSTM for time-series prediction: An application to weather forecasting},
  author={Karevan, Zahra and Suykens, Johan AK},
  journal={Neural Networks},
  volume={125},
  pages={1--9},
  year={2020},
  publisher={Elsevier}
}

@article{boussif2024improving,
  title={Improving* day-ahead* Solar Irradiance Time Series Forecasting by Leveraging Spatio-Temporal Context},
  author={Boussif, Oussama and Boukachab, Ghait and Assouline, Dan and Massaroli, Stefano and Yuan, Tianle and Benabbou, Loubna and Bengio, Yoshua},
  journal={Advances in Neural Information Processing Systems},
  volume={36},
  year={2024}
}

@article{novo2022planning,
  title={Planning the decarbonisation of energy systems: The importance of applying time series clustering to long-term models},
  author={Novo, Riccardo and Marocco, Paolo and Giorgi, Giuseppe and Lanzini, Andrea and Santarelli, Massimo and Mattiazzo, Giuliana},
  journal={Energy Conversion and Management: X},
  volume={15},
  pages={100274},
  year={2022},
  publisher={Elsevier}
}

@incollection{imambi2021pytorch,
  title={PyTorch},
  author={Imambi, Sagar and Prakash, Kolla Bhanu and Kanagachidambaresan, GR},
  booktitle={Programming with TensorFlow: solution for edge computing applications},
  pages={87--104},
  year={2021},
  publisher={Springer}
}

@article{chu2025sft,
  title={Sft memorizes, rl generalizes: A comparative study of foundation model post-training},
  author={Chu, Tianzhe and Zhai, Yuexiang and Yang, Jihan and Tong, Shengbang and Xie, Saining and Schuurmans, Dale and Le, Quoc V and Levine, Sergey and Ma, Yi},
  journal={arXiv preprint arXiv:2501.17161},
  year={2025}
}

@inproceedings{woo2024unified,
  title={Unified training of universal time series forecasting transformers},
  author={Woo, Gerald and Liu, Chenghao and Kumar, Akshat and Xiong, Caiming and Savarese, Silvio and Sahoo, Doyen},
  booktitle={Forty-first International Conference on Machine Learning},
  year={2024}
}

@article{liu2025empowering,
  title={Empowering time series analysis with synthetic data: A survey and outlook in the era of foundation models},
  author={Liu, Xu and Aksu, Taha and Liu, Juncheng and Wen, Qingsong and Liang, Yuxuan and Xiong, Caiming and Savarese, Silvio and Sahoo, Doyen and Li, Junnan and Liu, Chenghao},
  journal={arXiv preprint arXiv:2503.11411},
  year={2025}
}

@article{ye2026empowering,
  title={Empowering time series analysis with foundation models: A comprehensive survey},
  author={Ye, Jiexia and Yu, Yongzi and Zhang, Weiqi and Wang, Le and Li, Jia and Tsung, Fugee},
  journal={Information Fusion},
  pages={104601},
  year={2026},
  publisher={Elsevier}
}

@article{lara2020temporal,
  title={Temporal convolutional networks applied to energy-related time series forecasting},
  author={Lara-Ben{\'\i}tez, Pedro and Carranza-Garc{\'\i}a, Manuel and Luna-Romera, Jos{\'e} M and Riquelme, Jos{\'e} C},
  journal={applied sciences},
  volume={10},
  number={7},
  pages={2322},
  year={2020},
  publisher={MDPI}
}

@inproceedings{yoon2022robust,
  title={Robust probabilistic time series forecasting},
  author={Yoon, TaeHo and Park, Youngsuk and Ryu, Ernest K and Wang, Yuyang},
  booktitle={International Conference on Artificial Intelligence and Statistics},
  pages={1336--1358},
  year={2022},
  organization={PMLR}
}

@article{clark2004population,
  title={Population time series: process variability, observation errors, missing values, lags, and hidden states},
  author={Clark, James S and Bj{\o}rnstad, Ottar N},
  journal={Ecology},
  volume={85},
  number={11},
  pages={3140--3150},
  year={2004},
  publisher={Wiley Online Library}
}

@article{liu2026pre,
  title={From Pre-training to Post-training: A Survey on Time Series Foundation Models},
  author={Liu, Zhen and Li, Boyuan and Huang, Hao and Sun, Yanru and Wang, Yucheng and Wu, Min and Ma, Qianli},
  year={2026},
  publisher={TechRxiv}
}

@article{kottapalli2025foundation,
  title={Foundation models for time series: A survey},
  author={Kottapalli, Siva Rama Krishna and Hubli, Karthik and Chandrashekhara, Sandeep and Jain, Garima and Hubli, Sunayana and Botla, Gayathri and Doddaiah, Ramesh},
  journal={arXiv preprint arXiv:2504.04011},
  year={2025}
}

\end{document}